\pdfoutput=1

\documentclass[11pt]{article}

\usepackage{acl}

\usepackage{times}
\usepackage{latexsym}

\usepackage[T1]{fontenc}

\usepackage[utf8]{inputenc}

\usepackage{microtype}

\usepackage{inconsolata}

\usepackage{graphicx}
\usepackage{amsmath}       
\usepackage{CJKutf8}
\usepackage{array}
\usepackage{colortbl}  
\usepackage{booktabs}
\usepackage{pifont}
\usepackage{tcolorbox}
\tcbuselibrary{listingsutf8}
\usepackage{tikz}
\usepackage{changepage}
\usepackage{amssymb}
\usepackage{float}
\usepackage{graphicx}

\newcommand{\red}[1]{\textcolor{red}{#1}}
\newcommand{\midruleline}{
    \vspace{0.2em} 
    \noindent\begin{tikzpicture}
    \draw[dashed] (0,0) -- (1.0\linewidth,0);
    \end{tikzpicture}
    \vspace{0.4em}
}

\title{WebNovelBench: Placing LLM Novelists on the Web Novel Distribution}

\author {
    Leon Lin\textsuperscript{\rm 1},\qquad
    Jun Zheng\textsuperscript{\rm 2},\qquad
    Haidong Wang\textsuperscript{\rm 2}\\
    \textsuperscript{\rm 1}Nanyang Technological University, \textsuperscript{\rm 2}Sun Yat-Sen University \\
liangtao.lin@ntu.edu.sg, \{zhengj98, wanghd7\}@mail2.sysu.edu.cn\\\\
\includegraphics[height=1em]{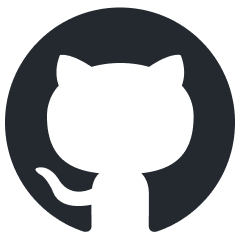} \url{https://github.com/OedonLestrange42/webnovelbench}\\
\includegraphics[height=1em]{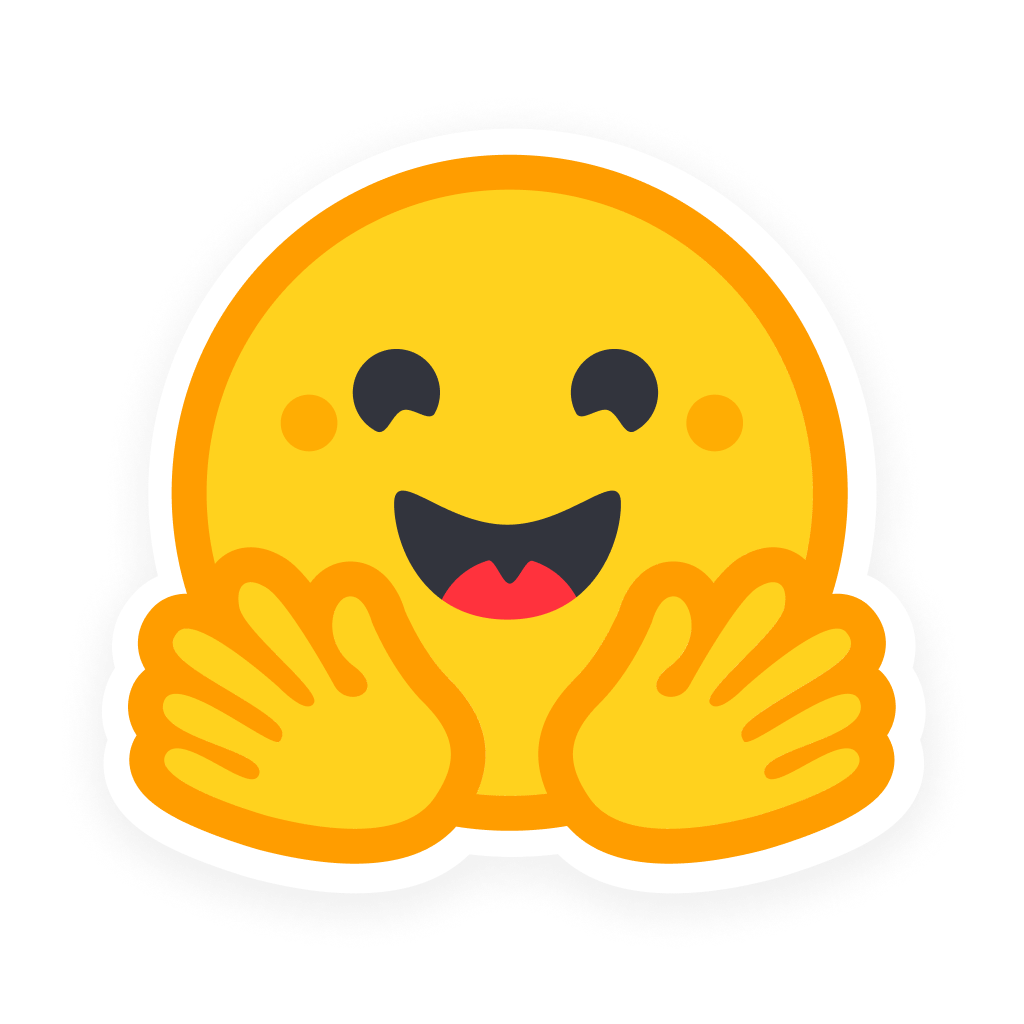} \url{https://huggingface.co/datasets/Oedon42/webnovelbench}\\
}

\begin{document}
\maketitle
\begin{abstract}
Robustly evaluating the long-form storytelling capabilities of Large Language Models (LLMs) remains a significant challenge, as existing benchmarks often lack the necessary scale, diversity, or objective measures. To address this, we introduce WebNovelBench, a novel benchmark specifically designed for evaluating long-form novel generation. WebNovelBench leverages a large-scale dataset of over 4,000 Chinese web novels, framing evaluation as a synopsis-to-story generation task. We propose a multi-faceted framework encompassing eight narrative quality dimensions, assessed automatically via an LLM-as-Judge approach. Scores are aggregated using Principal Component Analysis and mapped to a percentile rank against human-authored works. Our experiments demonstrate that WebNovelBench effectively differentiates between human-written masterpieces, popular web novels, and LLM-generated content. We provide a comprehensive analysis of 24 state-of-the-art LLMs, ranking their storytelling abilities and offering insights for future development. This benchmark provides a scalable, replicable, and data-driven methodology for assessing and advancing LLM-driven narrative generation.
\end{abstract}

\section{Introduction} 

Can Large Language Models (LLMs) generate stories that surpass human-written ones? Recent breakthroughs, exemplified by models like GPT-4o and Deepseek-R1~\citep{deepseekai2025deepseekr1incentivizingreasoningcapability}, underscore their remarkable ability to produce coherent, imaginative, and contextually nuanced narratives. This raises intriguing questions: How proficient are today's LLMs in story generation, and how do their outputs compare to human-authored works?

Evaluating LLM performance in this open-ended domain remains a significant challenge. While prior research has explored story generation evaluation~\citep{OpenMEVA,liu2024alignbenchbenchmarkingchinesealignment,paech2024eqbenchemotionalintelligencebenchmark,ismayilzada2025evaluatingcreativeshortstory}, these efforts often face limitations such as small dataset sizes or insufficient story diversity, hindering widespread adoption. This contrasts with fields like code generation and mathematical reasoning, where benchmarks such as CodeForces Rating~\citep{quan2025codeelo} and American Invitational Mathematics Examination 2024 (AIME 2024)~\citep{AIME2024} serve as widely accepted standards.

\begin{figure*}[t]
    \centering
    \includegraphics[width=1.\linewidth]{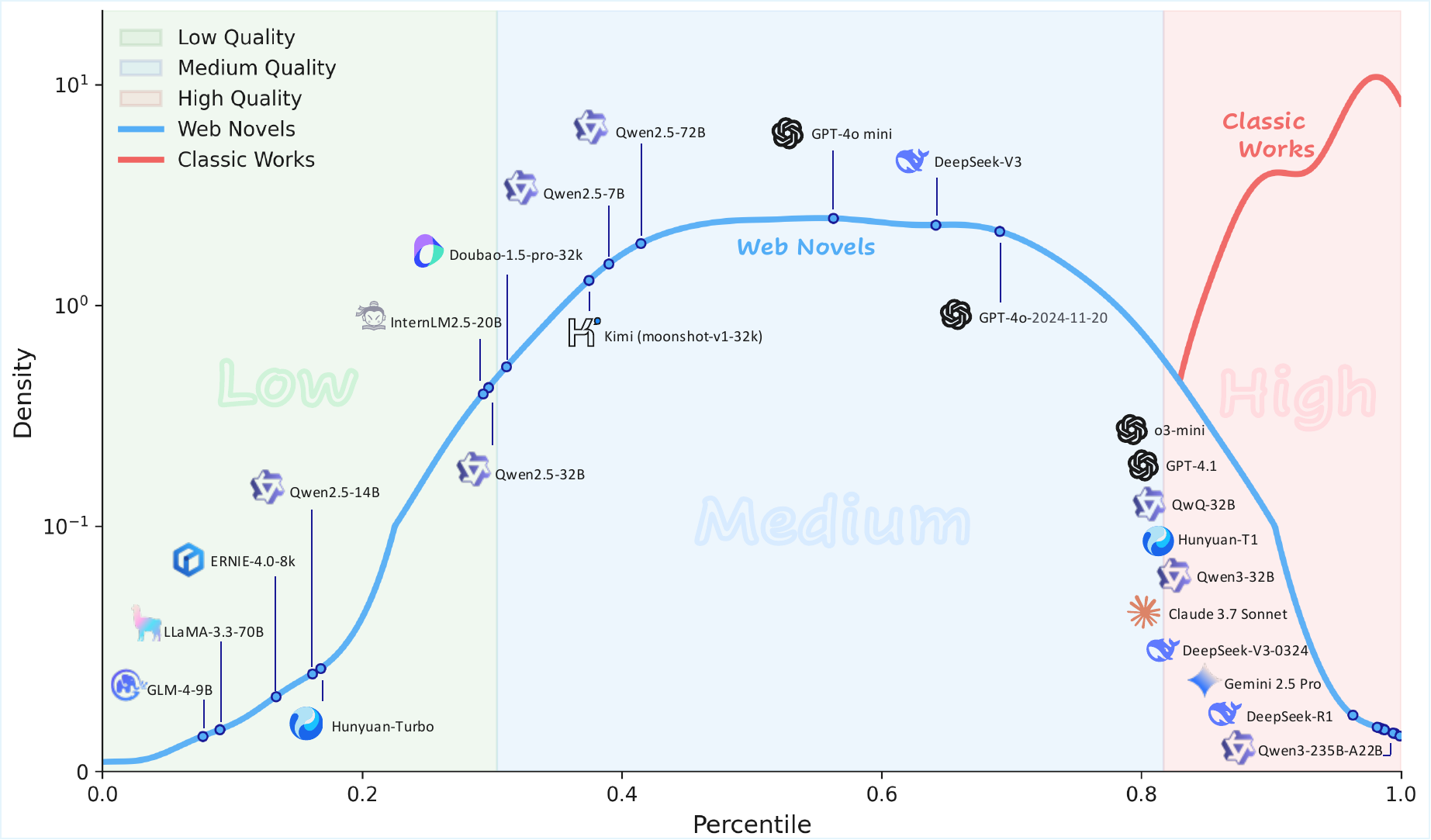}
    \caption{\textbf{Web Novel Dataset Distribution and LLM Placement. }
Our web novel dataset's quality distribution, with Low, Medium, and High zones (95\% CIs). The red curve (classic literary works) validates the high-quality zone. Positions of 24 LLMs indicate their performance relative to this corpus.}
    \label{fig:distribution}
\end{figure*}

Inspired by such successes, we propose WebNovelBench, a comprehensive and intuitive benchmark for story generation guided by three key principles:
\begin{itemize}
\item \textbf{Broad Data Foundation:} Utilizes diverse, popular human-authored works.
\item \textbf{Representative Tasks:} Covers diverse storytelling styles, themes, and complexities (details in Section~\ref{sec:dataset}).
\item \textbf{Automated and Objective Evaluation:} Minimizes subjectivity via robust, consistent automated methods.
\end{itemize}

We leverage 4,000+ popular Chinese web novels (>10,000 readers each) for a synopsis-to-story generation task. An LLM-as-Judge approach evaluates stories across eight narrative dimensions. Validation with Mao Dun Literature Prize novels, which scored high (Figure~\ref{fig:distribution}), confirms alignment with human judgment. WebNovelBench thus offers automatic assessment of LLM storytelling capabilities without extensive manual intervention, establishing a standardized framework for comparison against human-authored content. In summary, our main contributions are: 
\begin{itemize}
\item We introduce WebNovelBench, a large-scale, data-driven evaluation framework for story generation, accurately ranking human-authored and LLM-generated stories via distribution analysis.
\item We define eight evaluation dimensions for Chinese story quality, employing a validated LLM-as-Judge mechanism for robust automated evaluation.
\item We conduct a comprehensive evaluation of 24 state-of-the-art LLMs, ranking their storytelling abilities relative to human-authored works and offering insights for future development.
\end{itemize}




\section{Related Work}
\subsection{LLM General Benchmark} 


General LLM benchmarks like MMLU~\citep{MMLUhendryckstest2021} and its variants~\citep{wang2024mmlupro,yue2024mmmupro}, or dynamic benchmarks like MixEval~\citep{ni2024mixeval}, are invaluable for assessing broad capabilities such as reasoning and knowledge recall. However, they often lack the specific, nuanced criteria required to effectively evaluate creative, open-ended tasks like long-form story generation, particularly regarding narrative quality, coherence, and creativity. This highlights the need for specialized benchmarks in creative domains, akin to CodeForces Rating~\citep{quan2025codeelo} and SWE-Bench Verified~\citep{jimenez2024swebench} or AIME 2024~\citep{AIME2024} for coding and mathematics.

\subsection{Story Generation Benchmark}

Previous work on evaluating LLM-generated stories has explored creativity~\citep{ismayilzada2025evaluatingcreativeshortstory, paech2024eqbenchemotionalintelligencebenchmark} and the correlation of automatic metrics with human judgment~\citep{OpenMEVA}. For instance,~\citet{ismayilzada2025evaluatingcreativeshortstory} assessed LLM creativity in short story generation but used only four samples per LLM. EQ-Bench~\citep{paech2024eqbenchemotionalintelligencebenchmark} offers a creative writing score based on a limited set of thirty-two prompts. OpenMEVA~\citep{OpenMEVA} provided a framework for evaluating automatic metrics but did not focus on broad dataset diversity. These existing efforts often suffer from limitations such as small dataset sizes or insufficient story diversity, hindering the establishment of a universally accepted standard. Our work aims to address this gap by emphasizing a large, diverse dataset derived from widely-read web novels and a robust, automated evaluation protocol.

\subsection{LLM-as-a-Judge} 

Automated evaluation for open-ended text generation increasingly employs the LLM-as-a-Judge paradigm~\citep{zheng2023judgingllmasajudgemtbenchchatbot}, where LLMs assess outputs without reference texts~\citep{li2024prdpeerrankdiscussion, kasner2024traditionalbenchmarksanalyzingbehaviors}. While promising for aligning with human preferences, LLM-as-Judge reliability is a concern, with research focusing on fairness and potential biases~\citep{shi2025optimizationbasedpromptinjectionattack, zhang2023widerdeeperllmnetworks, ye2024justiceprejudicequantifyingbiases}. For instance, \citet{ye2024justiceprejudicequantifyingbiases} introduced CALM to identify and quantify biases in LLM judges. This body of work underscores the importance of careful design and validation of the LLM-as-Judge component, which we address in WebNovelBench to ensure reliable and fair story evaluation based on multi-dimensional criteria.

\section{Benchmark Construction}

\begin{figure*}[t]
\centering
\includegraphics[width=1.\linewidth]{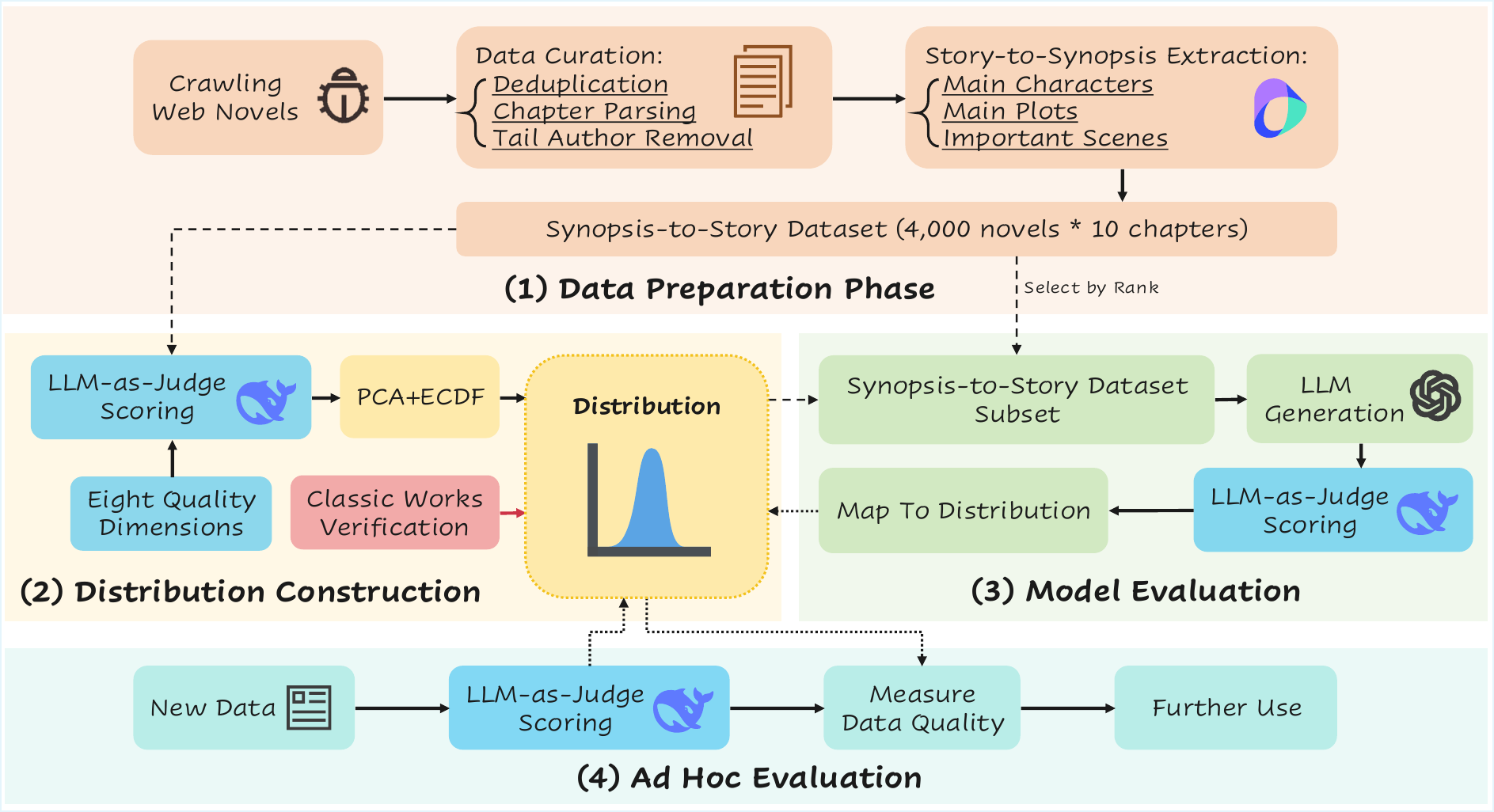}
\caption{\textbf{Framework of Our Method}.
Our benchmark framework consists of four major components:
(1) \textit{Data Preparation Phase:} We collect and curate a large web novel dataset, and use Doubao for story-to-synopsis extraction to build a 4,000 novels synopsis-to-story dataset.
(2) \textit{Distribution Construction:} Each story is scored across eight quality dimensions using LLM-as-judge, followed by PCA+ECDF to form a quality distribution benchmark. Classic literary works are used to validate the high end of the distribution.
(3) \textit{Model Evaluation:} LLMs generate stories from selected subsets of the dataset. Their outputs are scored and mapped onto the distribution to assess model performance.
(4) \textit{Ad Hoc Evaluation:} New data can be scored and aligned with the benchmark for measuring data quality and supporting further applications.}
\label{fig:Framework}
\end{figure*}

\subsection{Dataset} 
\label{sec:dataset}

We curated a dataset from over 10,000 Chinese web novels (published 2013-2020) via several preprocessing steps:

\begin{itemize}
    \item \textbf{Deduplication.} 
    Some text files may contain highly similar content under different titles, indicating that they represent the same novel. Directly comparing exact character matches is insufficient for identifying duplicates. To address this, we employed \texttt{difflib}\footnote{\url{https://docs.python.org/3/library/difflib.html}} to compute pairwise similarity scores and removed novels with a similarity score greater than 0.9.

    \item \textbf{Chapter Parsing.} 
    Web novels are typically long-form narratives presented in a serial-chapter format. However, the raw crawled texts lacked explicit chapter delimiters. To extract chapters, we designed regular expression patterns tailored to common chapter formatting styles in Chinese web novels. Novels containing fewer than ten chapters were excluded, as they did not meet our definition of “long-form” or suggested incomplete parsing.

    \item \textbf{Tail Author Removal.}
    It is commonly observed that successful authors tend to publish multiple novels over time. Based on this observation, we compiled the list of all authors and excluded novels written by those with the fewest works. After this filtering step, we retained a final set of 4,000 web novels.
\end{itemize}
The curated 4,000 novels ensure 'Representative Tasks' by covering diverse genres (e.g., Eastern Fantasy: 1281, General Realism: 1255, Western Fantasy: 670, Historical: 234, plus Sci-Fi, Suspense, Romance with varied subthemes like 'Original World Setting'). This breadth reflects popular web fiction complexities.

For the synopsis-to-story task, we utilize Doubao-pro-32k\footnote{\url{https://www.volcengine.com/product/doubao}} to generate a 'synopsis' (main characters, key plot points, important scenes) for ten random consecutive chapters from each novel. This yielded ten <chapter content, synopsis> pairs per novel (details in Appendix \ref{app:systemprompt}).

\subsection{Metric} 

We evaluate narrative quality using eight key dimensions (Table~\ref{table:metrics}) covering stylistic and structural elements (e.g., literary devices, character consistency). This provides a nuanced assessment beyond surface-level fluency.

\begin{table*}[htbp]
\centering
\renewcommand{\arraystretch}{1.3}
\resizebox{.9\textwidth}{!}{%
\begin{tabular}{|>{\raggedright\arraybackslash}p{3.4cm}
                |>{\raggedright\arraybackslash}p{2.8cm}
                |>{\raggedright\arraybackslash}p{7.0cm}
                |c|}
\hline
\textbf{Metric} & \begin{CJK*}{UTF8}{gbsn}\textbf{评估维度}\end{CJK*} & \textbf{Explanation} & \textbf{Weight} \\
\hline
D1: Use of Literary Devices & \begin{CJK*}{UTF8}{gbsn}修辞手法\end{CJK*} & Based on the quantity and quality of rhetorical devices like metaphor and symbolism & \underline{0.1304} \\
\hline
D2: Richness of Sensory Detail & \begin{CJK*}{UTF8}{gbsn}感官描述丰富度\end{CJK*} & Frequency of visual, auditory, and other sensory descriptions & 0.1160  \\
\hline
D3: Balance of Character Presence & \begin{CJK*}{UTF8}{gbsn}角色平衡度\end{CJK*} & Frequency, dialogue proportion, and psychological depth of each character & 0.1152  \\
\hline
D4: Distinctiveness of Character Dialogue & \begin{CJK*}{UTF8}{gbsn}角色对白独特性\end{CJK*} & Whether dialogue reflects distinct personalities & 0.1171 \\
\hline
D5: Consistency of Characterisation & \begin{CJK*}{UTF8}{gbsn}角色一致性\end{CJK*} & Whether language and actions align with character identity & \textbf{0.1377} \\
\hline
D6: Atmospheric and Thematic Alignment & \begin{CJK*}{UTF8}{gbsn}意境匹配度\end{CJK*} & Whether scenes support the overall atmosphere and themes & 0.1290  \\
\hline
D7: Contextual Appropriateness & \begin{CJK*}{UTF8}{gbsn}语境适配度\end{CJK*} & Whether settings match time/place/cultural background & 0.1281 \\
\hline
D8: Scene-to-Scene Coherence & \begin{CJK*}{UTF8}{gbsn}跨场景衔接度\end{CJK*} & Whether scene transitions are smooth and natural & 0.1263  \\
\hline
\end{tabular}%
}
\caption{\textbf{Narrative Evaluation Metrics and PCA-Derived Weights.}
This table lists the eight dimensions used to evaluate narrative quality. Each weight reflects the metric’s relative importance, derived through PCA on web novels scores.}
\label{table:metrics}
\end{table*}



\subsection{Scoring Method}

To rank LLM works against our 4,000 human-authored web novels, we combine Principal Component Analysis (PCA)~\citep{Hotelling1933AnalysisOA} for multi-dimensional score aggregation and Empirical Cumulative Distribution Function (ECDF)~\citep{conover1999practical} for percentile ranking.

\noindent
\textbf{Score Aggregation via PCA.} Given a dataset of $N$ samples, where each sample is evaluated along $d$ dimensions, we first normalize all dimension scores using z-score standardization: 

\begin{equation}
    z_{ij} = \frac{x_{ij} - \mu_j}{\sigma_j} (i\in[1, N], j\in[1,d]),
\end{equation}
where $x_{ij}$ is the raw score of the $i$-th sample in the $j$-th dimension, and $\mu_j$ and $\sigma_j$ are the mean and standard deviation of the $j$-th dimension across all samples.

We then perform PCA on the standardized data to extract the first principal component, whose normalized loading vector $\mathbf{w} = (w_1, w_2, \dots, w_d)$ represents the relative importance of each dimension. The aggregated composite score $s_i$ for the $i$-th sample is then computed as a weighted sum:

\begin{equation}
    s_i = \sum_{j=1}^{d} w_j z_{ij}
\end{equation}

\noindent
\textbf{Percentile Ranking via ECDF.} To translate raw scores into interpretable relative rankings, we apply the ECDF over all aggregated scores $\{s_1, s_2, \dots, s_N\}$:

\begin{equation}
    \mathrm{ECDF}(x) = \frac{1}{N} \sum_{i=1}^N \mathbb{I}(s_i \leq x),
\end{equation}


where $\mathbb{I}(\cdot)$ is the indicator function. The ECDF provides a percentile score in the range $[0, 1]$, representing the proportion of samples with scores less than or equal to $x$.

Given a new LLM-generated sample with aggregated score $s_{\text{new}}$, its percentile rank is $P_{\text{new}} = \mathrm{ECDF}(s_{\text{new}})$.


This percentile reflects the model's performance relative to the full empirical distribution of the reference dataset—indicating the level at which stories generated by a LLM are comparable to those written by humans. By evaluating the model across batches of test samples, we can estimate the LLM's overall writing ability.

Let \( \mathcal{B} = \{s^{(1)}_{\text{LLM}}, s^{(2)}_{\text{LLM}}, \dots, s^{(M)}_{\text{LLM}} \} \) be a set of aggregated scores for a batch of \( M \) LLM-generated samples. The estimated writing level is then defined as the average percentile:

\begin{equation}
    \hat{P}_{\text{LLM}} = \frac{1}{M} \sum_{m=1}^{M} \mathrm{ECDF}\left(s^{(m)}_{\text{LLM}}\right)
\end{equation}


The value \( \hat{P}_{\text{LLM}} \in [0, 1] \) represents the expected percentile rank of the LLM’s output relative to the distribution of human-written texts in the reference dataset.



\section{Experiments and Results}

\begin{figure*}[t]
\centering
\includegraphics[width=2.0\columnwidth]{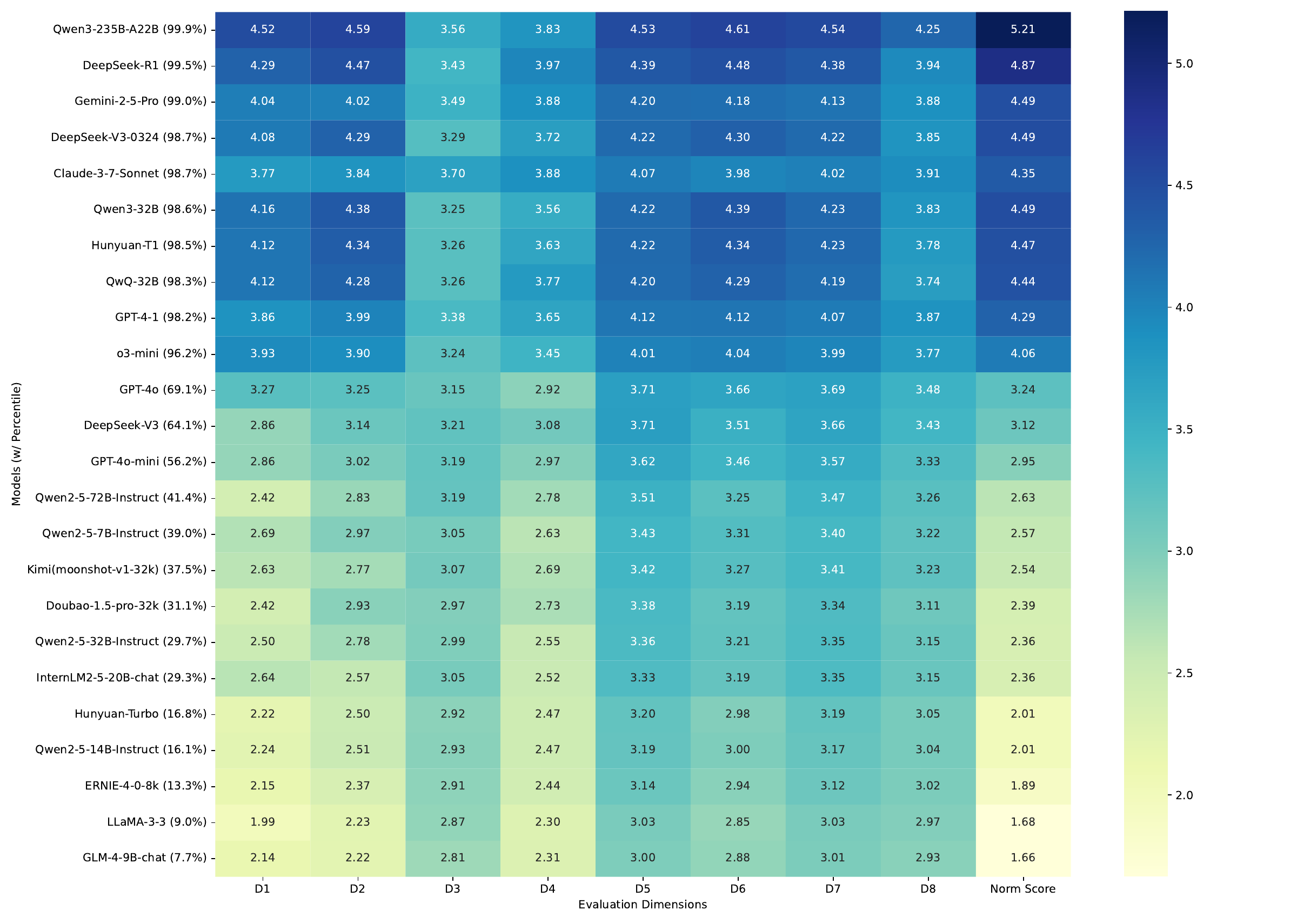}
\caption{\textbf{LLM Performance Heatmap Across Narrative Dimensions.}
Shows average scores (1-5 scale) for 24 LLMs on eight dimensions, sorted by percentile rank. Final column is PCA-derived weighted norm score. Higher scores indicate better alignment with quality human writing.}
\label{fig:heatmap}
\end{figure*}

\subsection{Experimental Setup}

Due to resource limits, our evaluation dataset uses 100 web novels (one per percentile from the 4,000-novel distribution), with 10 synopsis-to-story pairs per novel (1,000 test samples). An LLM's overall rank is its average percentile across these 100 books.

LLMs (open-source and proprietary) were accessed via APIs with a standardized system prompt and constant generation settings (4096 max tokens, temperature 0.6, eight evaluation criteria in prompt; see Appendix~\ref{app:systemprompt}). Outputs were evaluated by Deepseek-V3 using a consistent evaluation prompt. For details on the robustness analysis, see Section~\ref{sec:llm-as-judge} below.


\subsection{Main Experiments}
We evaluate a total of 13 open source models and 11 closed source models on our benchmark. Figure \ref{fig:heatmap} illustrates the performance of these frontier models across eight distinct narrative evaluation dimensions and overall effect.

Top models (Qwen3-235B-A22B~\citep{yang2025qwen3technicalreport}, DeepSeek-R1~\citep{deepseekai2025deepseekr1incentivizingreasoningcapability}, Gemini-2.5-Pro) score high (3.5-4.6) across dimensions; Qwen3-235B-A22B achieves a 5.21 norm score, indicating strong alignment with high-quality human writing. Mid-tier models (e.g., GPT-4o, DeepSeek-V3~\citep{deepseekai2025deepseekv3technicalreport}) show varied performance (scores 2.5-3.8), highlighting areas like sensory detail and literary device use for improvement. Lower-ranked models (e.g., GLM-4-9B-chat~\citep{glm2024chatglmfamilylargelanguage}, LLaMA-3-8B~\citep{grattafiori2024llama3herdmodels}) perform consistently poorly (less than 2.0 norm score), especially in literary devices and character dialogue, indicating significant room for improvement, particularly for open-source or smaller LLMs.



An intriguing observation from our analysis is the relatively narrow performance gap between top closed-source models (such as Claude-3-7-Sonnet and GPT-4.1) and leading open-source models (Qwen series and DeepSeek models), suggesting that open-source communities are rapidly bridging the performance divide traditionally held by proprietary models.

Overall, this benchmark effectively captures distinct strengths and weaknesses among current LLMs. While the most advanced models achieve near-perfect scores within our distribution, demonstrating their strong performance on the story-to-synopsis dataset, this does not diminish the value of the benchmark. On the contrary, it validates our original intuition. Our primary goal is to evaluate the story generation capabilities of contemporary LLMs, and the results suggest that leading models have reached a level comparable to top-tier works in web novels. This study primarily presents a methodological framework; for more fine-grained evaluation, future work may involve collecting higher-quality reference texts or designing more nuanced evaluation dimensions.




\begin{table*}[htbp]
\centering
\resizebox{1.0\textwidth}{!}{%
\begin{tabular}{l l l m{2cm} m{2cm} m{2cm} m{2cm}}
\toprule
\textbf{Benchmark} & \multicolumn{2}{c}{\textbf{Dataset Information}} & \multicolumn{3}{c}{\textbf{Evaluation Method}} \\
\cmidrule(lr){2-3} \cmidrule(lr){4-6}
 & Data Source & Testing Samples & Human Preference Alignment & Dimension Weights &  Evaluation Dimension \\
\midrule
OpenMEVA~\citep{OpenMEVA}  & Existing dataset    & 400 stories  & \ding{51}  & \ding{55}  & 8 \\
AlignBench Writing Ability~\citep{liu2024alignbenchbenchmarkingchinesealignment} & Self-constructed    & 75 stories & \ding{51}  & \ding{55} & 5 \\
EQ-Bench Longform Creative Writing~\citep{paech2024eqbenchemotionalintelligencebenchmark}   & Self-constructed  & 12 stories, each with 8 chapters & \ding{55}  & \ding{55} & 14 \\
\citet{ismayilzada2025evaluatingcreativeshortstory}  & Existing dataset & 4 stories & \ding{51} & \ding{55} & 4 \\
\textbf{WebNovelBench (ours)} &\textbf{ Web novels} & \textbf{100 stories, each with 10 chapters}  & \ding{51}  & \ding{51}   & \textbf{8} \\
\bottomrule
\end{tabular}%
}
\caption{Comparison of Other Benchmarks}
\label{tab:benchmarks}
\end{table*}

\subsection{Comparison with Other Benchmarks}
To contextualize our contribution, we compare WebNovelBench with existing benchmarks for story generation evaluation, as summarized in Table \ref{tab:benchmarks}.
OpenMEVA \citep{OpenMEVA} provides a framework for evaluating automatic metrics using an existing dataset of 400 stories. It aligns with human preference and utilizes 8 evaluation dimensions. However, it does not employ weighted dimensions for score aggregation, potentially treating all aspects of narrative quality as equally important, which might not reflect nuanced human judgment.
AlignBench Writing Ability \citep{liu2024alignbenchbenchmarkingchinesealignment} uses a self-constructed dataset of 75 stories. While it considers human preference and evaluates across 5 dimensions, its dataset size is relatively small, which might limit the diversity of narrative styles and scenarios covered. Similar to OpenMEVA, it does not incorporate dimension weights in its scoring.
EQ-Bench Longform Creative Writing \citep{paech2024eqbenchemotionalintelligencebenchmark} focuses on long-form creative writing with a self-constructed dataset of 12 stories, each comprising 8 chapters (totaling 96 evaluation instances). It employs a comprehensive set of 14 dimensions. However, according to the available information, it does not explicitly align its dataset construction with broad human preference (e.g., via popularity metrics of source texts) and also lacks a weighted approach to aggregating its numerous dimensional scores.
The work by \citet{ismayilzada2025evaluatingcreativeshortstory} evaluates creative short story generation using an existing dataset. While it aligns with human preference and uses 4 evaluation dimensions, its very small test set of only 4 stories per LLM significantly limits the robustness and generalizability of its findings. It also does not use weighted dimensions.
In contrast, WebNovelBench offers several key advantages:
\begin{itemize}
\item \textbf{Scale and Diversity:} Built on 4,000+ web novels, testing with 100 distinct stories (10 chapters each, 1,000 instances), ensuring wide genre/style coverage.
\item \textbf{Inherent Human Preference Alignment:} By using popular web novels (each with over 10,000 readers) as the source, our benchmark inherently captures broad human literary preferences.
\item \textbf{Data-Driven Dimension Weighting:} PCA derives weights for 8 narrative dimensions, providing a nuanced, objective assessment reflecting their relative importance in human-authored works.
\item \textbf{Comprehensive and Focused Evaluation:} Eight carefully defined dimensions provide a thorough yet focused assessment of key storytelling elements.
\end{itemize}
These features make WebNovelBench a robust, scalable, and replicable solution for assessing and advancing LLM-driven narrative generation, particularly for long-form stories in the Chinese web novel domain.

\section{Rationality Analysis}
\subsection{Metric Analysis}

To assess the rationality and effectiveness of the proposed eight evaluation metrics, we conducted a detailed statistical analysis combining principal component analysis (PCA) and distributional visualization.

\noindent
\textbf{Principal Component Analysis.} PCA shows the first component explains ~75.6\% of variance (first three greater than 90\%), indicating the metrics capture a dominant quality factor. Derived weights (11.5\%-13.8\%) are balanced, with 'Consistency of Characterisation' (Table~\ref{table:metrics}) highest, reflecting its discriminative power.\footnote{See Appendix \ref{app:pca} for further details.}

\begin{figure}[htbp!]
\centering
\includegraphics[width=1.0\columnwidth]{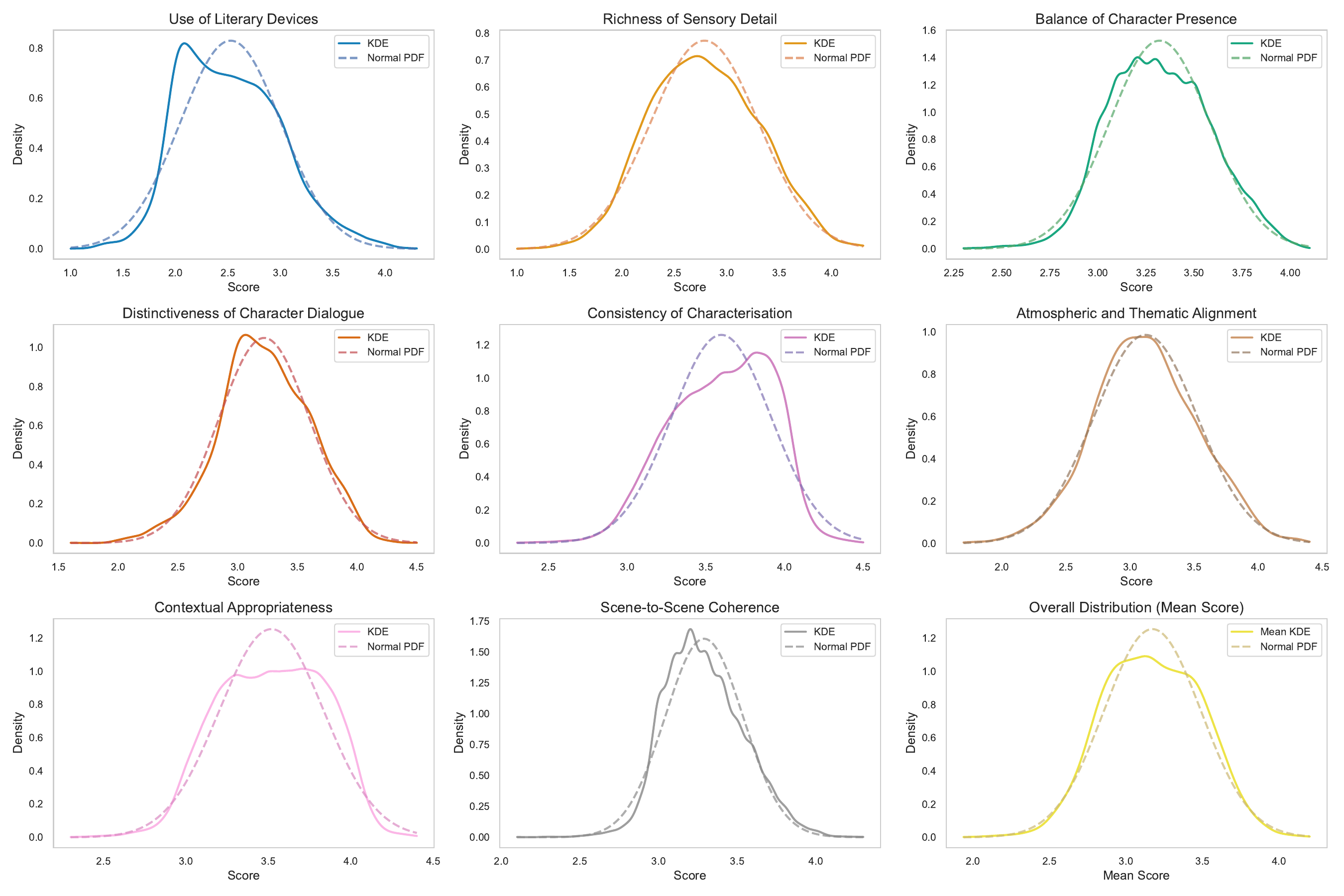}
\caption{\textbf{Distributions of Narrative Metrics and Fitted Normal Curves.}
Each subplot shows the empirical distribution (solid line) of a narrative evaluation dimension across the web novel dataset, alongside the corresponding fitted normal distribution (dashed line). The comparison illustrates the varying shapes of real-world data and highlights where distributions deviate from normality. The bottom-right panel presents the overall distribution of mean scores.}
\label{fig:metricsdistribution}
\end{figure}

\noindent
\textbf{Distributional Characteristics.}
The probability density function (PDF) of each individual metric, obtained via kernel density estimation (KDE)~\citep{parzen1962estimation}, is plotted against its best-fit normal distribution to evaluate shape characteristics (Figure \ref{fig:metricsdistribution}). The majority of metrics (e.g., D2, D3, D4, and D6) display near-Gaussian behavior, implying smooth and well-behaved scoring distributions conducive to comparative assessment. Metrics such as D1 (Use of Literary Devices) and D7 (Contextual Appropriateness) exhibit mild deviations from normality, with signs of skewness or slight multimodality. These deviations likely reflect the existence of content subgroups, for instance, differences in stylistic density between human and LLM-generated texts or variations in how explicitly context is embedded. Importantly, the averaged score across all dimensions yields an aggregate distribution that closely aligns with the Gaussian, further validating the integration of the metrics into a coherent composite score.

\noindent
\textbf{Implication.}
PCA and distributional analyses confirm our metrics are well-structured, diverse, complementary, and robust, suitable for large-scale evaluation of human and LLM narratives.

\subsection{Classic Literature Comparison}
We validated our benchmark against 25 Mao Dun Literature Prize-winning novels (first 10 chapters each). As shown in Figure~\ref{fig:distribution}, these classics consistently scored in the high range, confirming our framework's ability to capture acknowledged literary merit.

This comparative analysis not only aligns with human evaluative judgments but also affirms the anticipated quality hierarchy among the three text categories under study: classic works, web novels, and LLM-generated content. The findings validate the benchmark’s sensitivity to nuanced differences in textual quality and its robustness in reflecting the relative literary value of diverse sources. Moreover, the stratification of LLM-generated outputs into three distinct quality tiers based on this distribution appears both credible and well-justified.


\subsection{LLM-as-Judge}
\label{sec:llm-as-judge}
To eliminate human involvement in automatic evaluation, we adopt the LLM-as-Judge paradigm, employing Deepseek-V3 as the evaluator, which currently is one of the most advanced Chinese language models. Based on our intuition and empirical observations, models without explicit chain-of-thought reasoning tend to perform more efficiently and effectively on classification tasks of this nature. To mitigate position bias and context-length bias, issues shown to significantly impact pairwise comparison methods as demonstrated in \citet{ye2024justiceprejudicequantifyingbiases}, we adopt a direct scoring approach, where the LLM evaluates each generated output independently. This method not only reduces systematic biases but also enhances flexibility and scalability. Notably, it allows for the direct assessment of single-generation outputs, which is particularly valuable for tasks such as dataset cleaning and screening, critical processes in the development of high-quality LLMs.

\begin{figure}[htbp!]
\centering
\includegraphics[width=1.0\columnwidth]{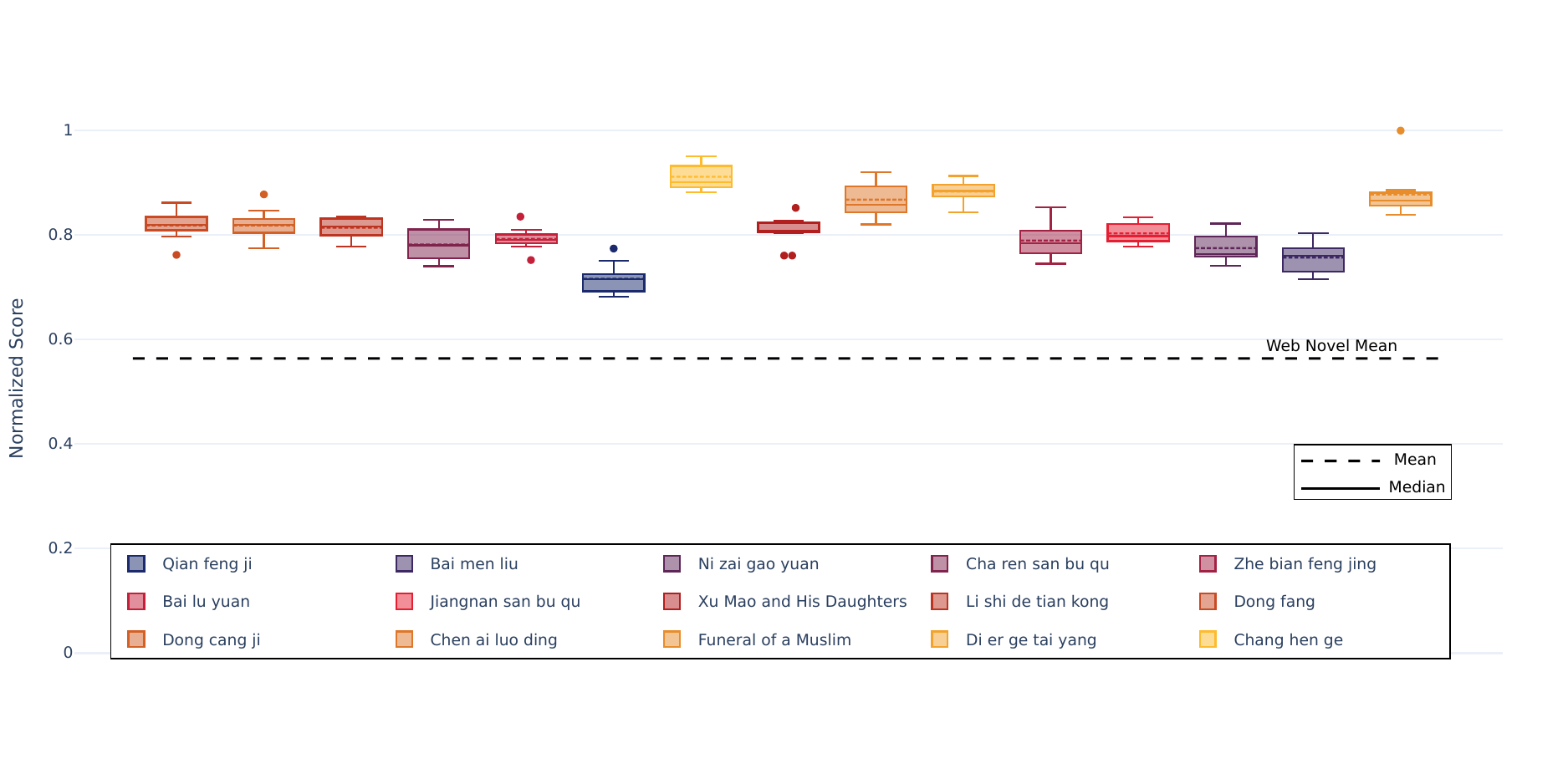}
\caption{\textbf{Robustness Assessment of LLM-as-Judge.} Boxplot of normalized scores for selected classic Chinese novels, based on 11 repeated evaluations using the LLM-as-judge framework. Each box shows the interquartile range (IQR) with the median (solid line) and mean (dashed line) marked. The majority of works demonstrate consistently high scores with narrow IQRs and minimal outliers, indicating the robustness and stability of model evaluations.}
\label{fig:llmasjudge}
\end{figure}

To demonstrate the robustness of the LLM-as-Judge approach, which is a concern of broad relevance, we conducted repeated experiments using a subset of classic works, which fall outside the distribution of our web novels benchmark. Specifically, we performed eleven independent evaluation rounds on the same dataset using identical Deepseek-V3 configurations. As shown in Figure \ref{fig:llmasjudge}, the boxplot exhibit high consistency, with an interquartile range (IQR) below 0.05 and a score variance within the 0.001 range, confirming the stability of both the model and the evaluation prompt design. These findings reinforce the reliability and credibility of our benchmark framework.

\subsection{Length Analysis}
To avoid introducing bias, we conducted an analysis of the length of model-generated outputs. The results indicate that most models closely adhered to the requested length or context window, producing outputs averaging between 800 and 1200 words. Notable exceptions include Claude 3.7 Sonnet and Gemini 2.5 Pro, which consistently generated significantly longer texts. Overall, output length remained relatively stable across models and did not emerge as a major differentiating factor under the 4096-token constraint. In future work, a scoring regularization term based on output length may be introduced to enhance robustness.\footnote{See Appendix \ref{app:length} for further details.}

\section{Conclusion}

WebNovelBench addresses challenges in evaluating LLM long-form storytelling. It uses 4,000+ Chinese web novels for a synopsis-to-story task. An automated pipeline with eight LLM-judged narrative dimensions and PCA+ECDF scoring provides percentile rankings against human content. Experiments show WebNovelBench effectively distinguishes classic literature, web fiction, and LLM outputs, providing stable rankings for 24 SOTA LLMs. WebNovelBench is a valuable tool for benchmarking progress and guiding LLM development in creative narrative generation. While focused on Chinese web novels, its principles are extendable. Future work includes diverse judge models, genre expansion, and fostering more engaging LLM storytellers, catalyzing innovation in machine-generated narratives.

\section*{Limitations}
Here we outline several limitations of our work. First, our benchmark relies exclusively on Chinese web novels as the evaluation dataset. While this provides a rich, diverse, and representative corpus for our purposes, future work should extend the benchmark to other languages and literary forms to improve its generalizability. Second, due to resource and time constraints, our experimental scale is limited: we evaluated performance only on the subset using a single LLM-as-Judge model. Although our results demonstrate robustness, evaluating additional subsets with multiple judge models in future studies would further strengthen and validate our conclusions. Lastly, while we propose that our benchmark can directly assess data quality, we have not yet explored its broader applications. Future research will investigate how these benchmark datasets can be leveraged to enhance model performance and other downstream tasks.

\bibliography{custom}

\clearpage
\appendix



\section{System Prompt}
\label{app:systemprompt}
To ensure consistency and clarity in our evaluation pipeline, we standardised the prompts used across all stages of the benchmark. Figure~\ref{fig:gen-system-prompt} shows the system prompt used to guide story generation, while Figure~\ref{fig:eval-system-prompt} presents the evaluation prompt employed by the LLM-as-judge to assess narrative quality. Additionally, Figure~\ref{fig:extract-system-prompt} illustrates the system prompt used for Story-to-Synopsis extraction, which serves as a foundation for constructing our generation dataset. These prompts were carefully designed to minimise ambiguity and ensure the reproducibility of our benchmark results.

\section{Data Samples}
\label{app:datasample}

Story-to-Synopsis Extraction Sample, generated by Doubao-pro-32k:
\begin{figure}[H]
\centering
\begin{tcolorbox}[colback=olive!5, colframe=olive!60,
                  width=0.95\linewidth,
                  title=Story-to-Synopsis Extraction Input,
                  fonttitle=\bfseries,
                  listing only,
                  listing options={
                      basicstyle=\ttfamily\footnotesize,
                      breaklines=true,
                      language=,
                      escapeinside=||,
                  }]
\begin{CJK*}{UTF8}{gbsn}{第1498章 不知好歹“萧炎哥哥，这药万归乃是四星初期的斗圣强者，实力极强，对付起来可并不容易。” ...... “废物一般的东西，也敢在我老师面前叫嚣，这一巴掌，就当是个小小教训！”}\end{CJK*}\\
\end{tcolorbox}
\end{figure}

\begin{figure}[H]
\centering
\begin{tcolorbox}[colback=olive!5, colframe=olive!60,
                  width=0.95\linewidth,
                  title=Story-to-Synopsis Extraction Output,
                  fonttitle=\bfseries,
                  listing only,
                  listing options={
                      basicstyle=\ttfamily\footnotesize,
                      breaklines=true,
                      language=,
                      escapeinside=||,
                  }]
\begin{CJK*}{UTF8}{gbsn}{<主要人物> 萧炎. 薰儿, 药万归, 药老, 药天 </主要人物> \\<主要情节> (1) 薰儿提醒萧炎药万归是四星初期斗圣强者，对付不易，萧炎表示自有分寸，只要得到并炼化净莲妖火，灭杀药万归不在话下。 ...... (5) 药天愤怒让药万归杀萧炎，萧炎灵魂分身出现，打了药天一巴掌并教训他。 </主要情节> \\<重要场景> 天空（众人聚集在此发生冲突的地点） </重要场景>}\end{CJK*}\\
\end{tcolorbox}
\end{figure}

LLMs synopsis-to-story generation samples, using the above extraction result as input example:
\begin{figure}[H]
\centering
\begin{tcolorbox}[colback=red!5, colframe=red!60,
                  width=0.95\linewidth,
                  title=Qwen3-235B-A22B Generation,
                  fonttitle=\bfseries,
                  listing only,
                  listing options={
                      basicstyle=\ttfamily\footnotesize,
                      breaklines=true,
                      language=,
                      escapeinside=||,
                  }]
\begin{CJK*}{UTF8}{gbsn}{浮空岛的云海翻涌如沸腾的熔岩，萧炎负手立于青石栏前，玄衣翻卷间露出腰间漆黑的陨落心炎。薰儿忽然按住他的手腕，指尖微颤，"那老者眉心的七色丹纹..."少女的呢喃被罡风撕碎，却在萧炎耳畔凝成清晰的警告："药族三长老药万归，四星斗圣的威压能压弯古树年轮。" ...... 药天的衣袍被狂风撕裂，露出心口狰狞的"罪丹"疤痕。分身降临的萧炎踩着虚空踏出八道残影，每记响亮的耳光都让云层塌陷成莲瓣："你可知你祖父当年偷炼'化骨丹'时，药老替他背下了多少罪行？"最后一掌落下时，浮空岛深处传来妖火的长吟，净莲虚影在所有人额头绽开血色印记。}\end{CJK*}\\
\end{tcolorbox}
\end{figure}

\begin{figure}[H]
\centering
\begin{tcolorbox}[colback=red!5, colframe=red!60,
                  width=0.95\linewidth,
                  title=GPT-4.1 Generation,
                  fonttitle=\bfseries,
                  listing only,
                  listing options={
                      basicstyle=\ttfamily\footnotesize,
                      breaklines=true,
                      language=,
                      escapeinside=||,
                  }]
\begin{CJK*}{UTF8}{gbsn}{天空如同被烈焰炙烤的铁幕，高远而幽深，浮云仿佛被无形之力撕裂成碎絮，斑驳地掩映着一场即将爆发的风暴。浩渺的天际下，人影浮现，气息如潮水般搅动空气，斗气波澜层层激荡。薰儿身着淡青长裙，立于萧炎身侧，眉宇间藏着一抹忧色，她低声道：“萧炎哥哥，药万归乃四星初期斗圣强者，不可大意。”她声音纤细，却带着隐隐的紧迫，如同清泉流淌过心头石隙。萧炎目光沉静如潭水，嘴角勾起自信的弧度。“放心，只要夺得并炼化净莲妖火，区区药万归，自然不在话下。” ...... 药天面色铁青，怒吼道：“杀了他！我要亲手取他性命！” 话音未落，虚空骤然扭曲，一道灵魂分身悄然凝聚于药天身侧。只听“啪”的一声脆响——灵魂分身挥手给了药天一个响亮耳光。天地间仿佛静止了一瞬，只剩下风声和众人的倒吸冷气。萧炎分身俯瞰着愣神的药天，语气冰冷：“自以为是，只会辱骂前辈，这就是你所谓的尊严？若再挑衅，我让你永远闭嘴。”}\end{CJK*}\\
\end{tcolorbox}
\end{figure}

\begin{figure}[H]
\centering
\begin{tcolorbox}[colback=red!5, colframe=red!60,
                  width=0.95\linewidth,
                  title=GLM-4-9B Generation,
                  fonttitle=\bfseries,
                  listing only,
                  listing options={
                      basicstyle=\ttfamily\footnotesize,
                      breaklines=true,
                      language=,
                      escapeinside=||,
                  }]
\begin{CJK*}{UTF8}{gbsn}{在广袤的天空之下，人声鼎沸，众人的目光都被那三道身影牢牢锁定。薰儿紧紧地站在萧炎身旁，她的眼中闪烁着坚定与担忧，仿佛能感受到萧炎内心的决心。“药万归，你不是四星初期斗圣吗？对付你，我萧炎自有分寸。”萧炎的声音在空中回荡，带着一丝不容置疑的威严。 ...... 话音刚落，药天便释放出强大的灵魂力量，向萧炎发动攻击。然而，萧炎的反应却出乎所有人的意料，他轻易地捏碎了药天的灵魂力量。“你！”药天愤怒地看向药万归，“杀了他！”萧炎的灵魂分身出现在众人面前，他毫不犹豫地给了药天一巴掌：“教训你一顿。”}\end{CJK*}\\
\end{tcolorbox}
\end{figure}

LLM-as-Judge evaluation sample, using Deepseek-V3 as a critic:

\begin{figure}[H]
\centering
\begin{tcolorbox}[colback=blue!5, colframe=blue!60,
                  width=0.95\linewidth,
                  title=Deepseek-V3 Evaluation,
                  fonttitle=\bfseries,
                  listing only,
                  listing options={
                      basicstyle=\ttfamily\footnotesize,
                      breaklines=true,
                      language=,
                      escapeinside=||,
                  }]
\begin{CJK*}{UTF8}{gbsn}{<提取结果>  <主要人物及其对白>  1. **萧炎**     - “放心，只要夺得并炼化净莲妖火，区区药万归，自然不在话下。”...  1. **场景铺垫**：天空被烈焰炙烤，风暴即将爆发，萧炎与薰儿在紧张的氛围中对话，薰儿提醒萧炎药万归的强大。 ... </主要情节>  <重要场景>  1. **天空下的对峙场景**：天空被烈焰炙烤，云海翻滚，萧炎与药万归双方针锋相对。 ... </重要场景>  </提取结果>  \\\\<评分结果>  <修辞手法评分>4</修辞手法评分>  <感官描述丰富度评分>4</感官描述丰富度评分>  <角色平衡度评分>3</角色平衡度评分>  <角色对白独特性评分>4</角色对白独特性评分>  <角色一致性评分>4</角色一致性评分>  <意境匹配度评分>4</意境匹配度评分>  <语境适配度评分>4</语境适配度评分>  <跨场景衔接度评分>4</跨场景衔接度评分>  </评分结果>}\end{CJK*}\\
\end{tcolorbox}
\end{figure}

\section{Ethical Considerations}
\label{app:ethical}
Our WebNovelBench dataset comprises over 4,000 publicly available Chinese web novels freely sourced from the internet. For each novel, we extract only 10 chapters, which is only a very small fraction of the full content, to minimize data usage while enabling robust evaluation. These texts contain no personally identifiable information or private data and are used solely for statistical evaluation and benchmarking purposes. They are not employed for model training or direct fine-tuning.

We strictly adhere to the license or terms for use of each LLM during inference testing. The models generate scientific artefacts based on our synopsis-to-story dataset, and these outputs are used exclusively for research evaluation and capability assessment. No generated content is repurposed for other uses.

Our use of these materials aligns with fair use principles for research. We conducted checks to ensure no offensive or sensitive personal content is included. While our benchmark specifically addresses Chinese web novels, potentially limiting generalizability, it poses minimal risks related to misuse, privacy, or bias.

\section{Principal Component Analysis Details}
\label{app:pca}

\begin{figure}[htbp!]
  \centering
  \includegraphics[width=1.0\linewidth]{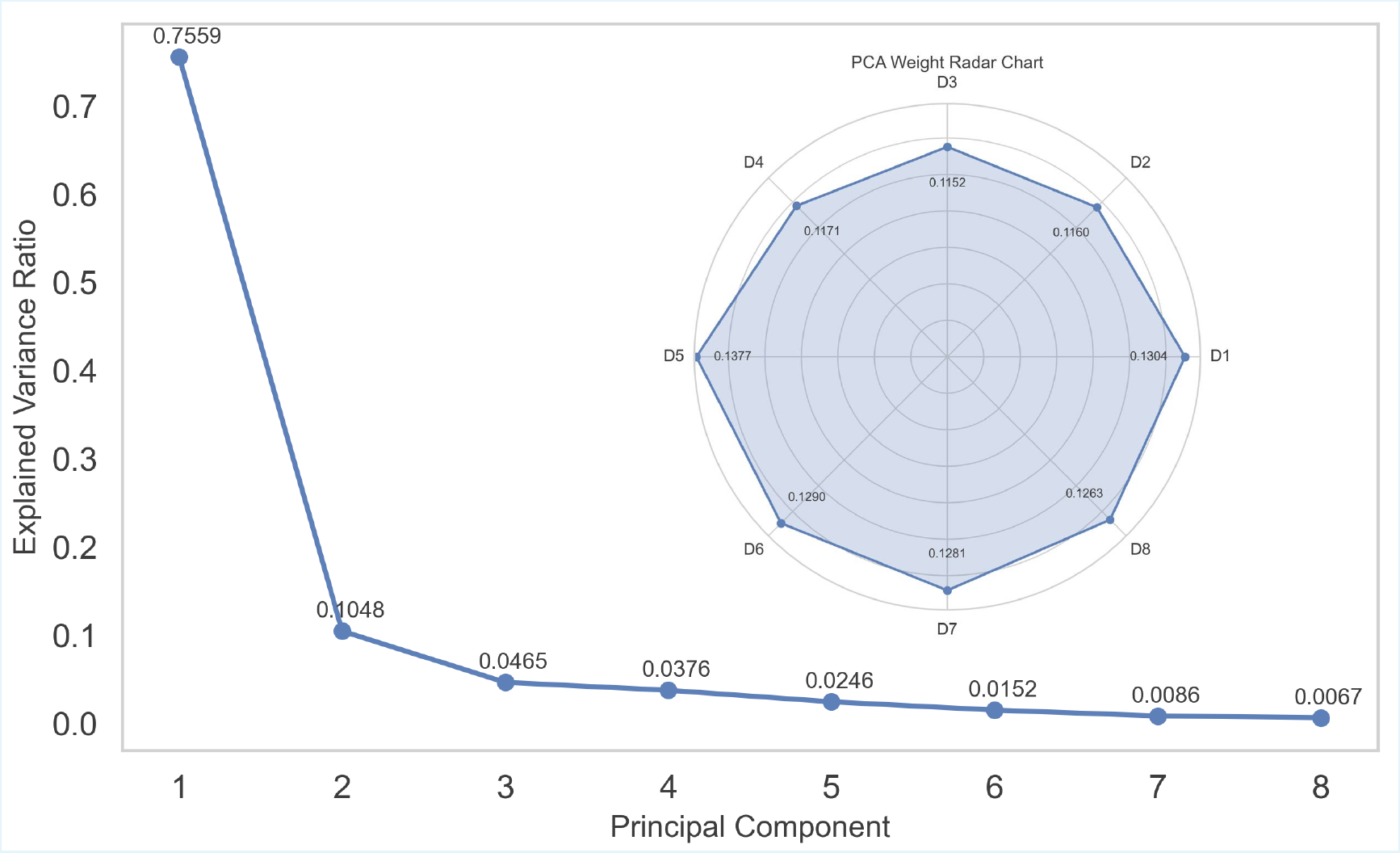}
  \caption{\textbf{PCA analysis of evaluation metrics.} The bar chart shows the explained variance ratio for each principal component. The radar chart visualises the relative weights of the eight narrative dimensions used in our benchmark.}
  \label{fig:pca_details}
\end{figure}

To determine the relative importance of each evaluation dimension, we applied Principal Component Analysis (PCA) to the score distributions across our web novel dataset. The scree plot in Figure~\ref{fig:pca_details} shows that the first principal component accounts for over 75\% of the total variance. This indicates that while each of our eight evaluation dimensions captures a distinct and meaningful aspect of narrative quality, they also collectively reflect a strong underlying evaluative signal. The high explained variance supports the internal coherence of our metric design and justifies the use of PCA-derived weights for aggregating narrative quality scores. This balance suggests that the dimensions are complementary rather than redundant, each contributing uniquely to the overall narrative assessment.

The radar chart embedded in the same figure visualises the PCA-derived weights assigned to each of the eight narrative dimensions (D1–D8). These weights, used throughout our benchmark scoring, reflect each dimension’s contribution to the primary variance component and therefore represent their relative importance in the overall evaluation framework.

\section{Length Analysis Details}
\label{app:length}

\begin{figure}[htbp!]
  \centering
  \includegraphics[width=\linewidth]{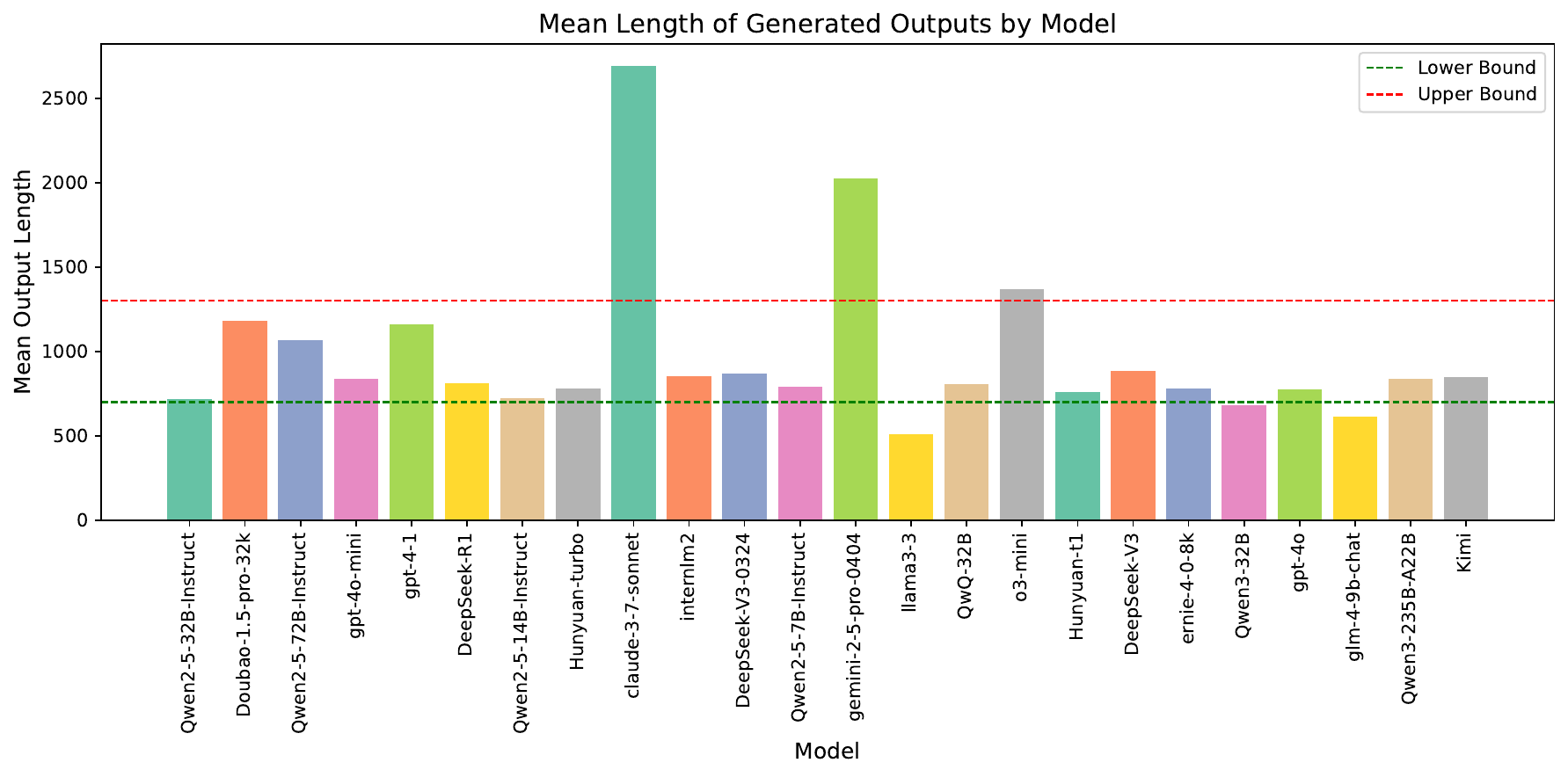}
  \caption{\textbf{Mean length of generated outputs by model.}}
  \label{fig:length_analysis}
\end{figure}

To support the main findings in Section~5.4, we provide a visual summary of the mean output lengths across all evaluated models in Figure~\ref{fig:length_analysis}. The green and red dashed lines represent the expected length bounds (800–1200 words). As shown, the vast majority of models generated outputs that fall within or near this range, indicating consistent adherence to the specified context length.

Given this overall consistency, we do not delve into detailed length-based comparisons in the main text. Notable outliers such as Claude 3.7 Sonnet (around 2,700) and Gemini 2.5 Pro produced (around 2,000) significantly longer outputs, while models like LLaMA 3.3 and GLM-4-9B-chat tended to under-generate. These deviations are exceptions rather than the norm and had limited impact on the overall evaluation results.

While length was not found to be a major differentiating factor in narrative quality, future iterations of the benchmark may consider applying soft constraints or regularisation mechanisms to penalise excessively long or short outputs.

\begin{figure*}[htbp!]
\centering
\begin{tcolorbox}[colback=gray!5, colframe=gray!60,
                  width=0.95\linewidth,
                  title=Story-to-Synopsis Extraction System Prompt,
                  fonttitle=\bfseries,
                  listing only,
                  listing options={
                      basicstyle=\ttfamily\footnotesize,
                      breaklines=true,
                      language=,
                      escapeinside=||,
                  }]
\begin{CJK*}{UTF8}{gbsn}{你的任务是从给定的小说片段中提取主要人物、情节和场景等信息，以便生成该小说的知识图谱和百科信息。请仔细阅读以下小说文本：\\
<小说> \red{\{text\}} </小说> \\
在提取信息时，请遵循以下步骤：\\
1. 仔细通读整个小说文本。 \\
2. 识别出主要人物，主要人物是在小说中起到关键作用、有较多情节围绕的角色。\\
3. 梳理主要情节，主要情节是推动故事发展的核心事件和关键转折。 \\
4. 确定重要场景，重要场景是故事发生的关键地点和环境。 \\
5. 检查提取的信息是否准确和完整。 \\
请在<提取结果>标签内输出你的提取结果，格式如下：\\
<主要人物> [列出主要人物的名字，并且用逗号分隔] </主要人物> \\
<主要情节> [详细描述主要情节，按照事件发展顺序。如(1)...(2)...，情节之间使用换行符分隔] </主要情节> \\
<重要场景> [列出重要场景的名称，并且用逗号分隔] </重要场景> \\
请确保提取的信息丰富、全面且准确。}\end{CJK*}\\
\midruleline
\\Your task is to extract key information—such as main characters, major plot points, and important settings—from the given novel excerpt. This information will be used to construct a knowledge graph and encyclopaedic entry for the novel. Please read the following text carefully:

<Novel> \red{\{text\}} </Novel>\\\\
When extracting the information, follow these steps:

1. Carefully read through the entire novel excerpt.

2. Identify the main characters, i.e., the characters who play a central role and around whom significant parts of the plot revolve.

3. Outline the major plot points, which refer to the core events and pivotal turns that drive the story forward.

4. Determine the important settings, i.e., the key locations and environments where significant story developments occur.

5. Check the extracted information for accuracy and completeness.\\\\
Output your extracted results within the <Extraction> tags using the following format:

<Main Characters> [List the names of the main characters, separated by commas] </Main Characters> 

<Main Plots> [Describe the main plot points in detail, following the chronological order of events. Use line breaks between different events, e.g., (1)...(2)...] </Main Plots> 

<Important Scenes> [List the names of the important settings, separated by commas] </Important Scenes>\\\\
Please ensure that the extracted information is rich, comprehensive, and accurate.

\end{tcolorbox}
\caption{System Prompt Used for Story-to-Synopsis Extraction}
\label{fig:extract-system-prompt}
\end{figure*}

\begin{figure*}[htbp!]
\centering
\begin{tcolorbox}[colback=gray!5, colframe=gray!60,
                  width=0.95\linewidth,
                  title=Generation System Prompt,
                  fonttitle=\bfseries,
                  listing only,
                  listing options={
                      basicstyle=\ttfamily\footnotesize,
                      breaklines=true,
                      language=,
                      escapeinside=||,
                  }]
\begin{CJK*}{UTF8}{gbsn}{你是一个中文小说作家，你需要根据用户提供的信息进行扩写创作，创作需要满足下列条件：\\
1. 用户会用下面的格式给出长篇小说的主要人物、主要情节和主要场景，请仔细阅读用户提供的信息：\\
<主要人物>[主要人物的名字]</主要人物>\\ 
<主要情节>[主要情节，按照事件发展顺序]</主要情节> \\
<重要场景>[重要场景的名称]</重要场景>\\
2. 评论家会根据下列标准打分：\\
    根据复杂修辞（隐喻/象征/悖论）的数量与质量提炼度，对修辞手法评分\\ 
    根据文本中的视觉、听觉、嗅觉等描写数量，对感官描述丰富度评分\\  
    统计每个角色在生成内容中的出现频率、对话占比、心理描写和评估人物描述的平衡度，对角色平衡度评分\\ 
    查看角色台词是否能反映本身个性，遮住名字后是否有区分度，对角色对白独特性评分\\ 
    分析角色语言、动作是否匹配其身份和背景，对角色一致性评分\\ 
    通过情感色谱分析，检查场景描写是否服务于整体氛围，对意境匹配度评分\\ 
    通过分析环境细节是否适应时代/地域背景，对语境适配度评分\\ 
    评估生成内容是否自然衔接不同场景从而避免场景割裂，对跨场景衔接度评分\\
3. 只需按照指定格式返回生成的小说：\\
    <text>你生成的小说内容</text>}\end{CJK*}\\
\midruleline
\\You are a Chinese fiction writer. Your task is to expand and create a narrative based on the information provided by the user. Your writing must adhere to the following guidelines:\\
1. The user will provide the key information for a long-form novel using the following format. Please read the information carefully:\\
<Main Characters>[Names of the main characters]</Main Characters>\\
<Main Plots>[Main plot points in chronological order]</Main Plots>\\
<Important Scenes>[Names of important scenes or locations]</Important Scenes> \\
2. A critic will evaluate your writing according to the following criteria:\\
Use of Literary Devices: Scored based on the quantity and refinement of complex rhetorical devices such as metaphor, symbolism, and paradox.\\
Richness of Sensory Detail: Scored based on the frequency of visual, auditory, olfactory, and other sensory descriptions.\\
Balance of Character Presence: Scored based on the frequency of each character’s appearance, proportion of dialogue, psychological depiction, and overall balance of character portrayal.\\
Distinctiveness of Character Dialogue: Scored based on whether each character’s dialogue reflects individual personality and remains distinguishable even if names are hidden.\\
Consistency of Characterisation: Scored based on whether the characters’ language and actions align with their identities and backgrounds.\\
Atmospheric and Thematic Alignment: Scored based on whether scene descriptions support the emotional tone and thematic coherence of the narrative.\\
Contextual Appropriateness: Scored based on whether the setting details are appropriate for the time period and regional background.\\
Scene-to-Scene Coherence: Scored based on whether the narrative transitions naturally between scenes, avoiding abrupt or disjointed shifts.\\
3. Return only the generated novel in the following format:\\
<text>Your generated story content</text>

\end{tcolorbox}
\caption{System Prompt Used for Generation}
\label{fig:gen-system-prompt}
\end{figure*}

\begin{figure*}[htbp!]
\centering
\begin{tcolorbox}[colback=gray!5, colframe=gray!60,
                  width=0.95\linewidth,
                  title=Evaluation System Prompt,
                  fonttitle=\bfseries,
                  listing only,
                  listing options={
                      basicstyle=\ttfamily\footnotesize,
                      breaklines=true,
                      language=,
                      escapeinside=||,
                  }]
\begin{CJK*}{UTF8}{gbsn}{你的任务是根据给定的指标规则对小说进行评分(1-5)。请仔细阅读以下小说文本：<小说> \red{\{chapter\}} </小说> 在提取信息时，请遵循以下步骤：\\
1. 仔细通读整个小说文本 \\
2. 识别出主要人物，主要人物是在小说中起到关键作用、有较多情节围绕的角色 \\
3. 梳理主要情节，主要情节是推动故事发展的核心事件和关键转折 \\
4. 确定重要场景，重要场景是故事发生的关键地点和环境 \\
5. 检查提取的信息是否准确和完整\\\\  
请在<提取结果>标签内输出你的提取结果，格式如下：\\
<主要人物及其对白>[列出主要人物的名字和对白]</主要人物及其对白> \\
<主要情节>[详细描述主要情节，按照事件发展顺序]</主要情节> \\
<重要场景>[列出重要场景的名称]</重要场景> \\\\
请确保提取的信息丰富、全面且准确。在评分时，请遵循以下步骤：\\
1. 根据复杂修辞（隐喻/象征/悖论）的数量与质量提炼度，给出修辞手法评分 \\
2. 根据文本中的视觉、听觉、嗅觉等描写数量，给出感官描述丰富度评分 \\
3. 统计每个角色在生成内容中的出现频率、对话占比、心理描写和评估人物描述的平衡度，给出角色平衡度评分 \\
4. 查看角色台词是否能反映本身个性，遮住名字后是否有区分度，给出角色对白独特性评分 \\
5. 分析角色语言、动作是否匹配其身份和背景，给出角色一致性评分 \\
6. 通过情感色谱分析，检查场景描写是否服务于整体氛围，给出意境匹配度评分 \\
7. 通过分析环境细节是否适应时代/地域背景，给出语境适配度评分 \\
8. 评估生成内容是否自然衔接不同场景从而避免场景割裂，给出跨场景衔接度评分。\\\\
请在<评分结果>标签内输出你的评分结果，格式如下：\\
<修辞手法评分>1</修辞手法评分> <感官描述丰富度评分>1</感官描述丰富度评分> <角色平衡度评分>1</角色平衡度评分> <角色对白独特性评分>1</角色对白独特性评分> <角色一致性评分>1</角色一致性评分> <意境匹配度评分>1</意境匹配度评分> <语境适配度评分>1</语境适配度评分> <跨场景衔接度评分>1</跨场景衔接度评分>  \\\\请确保评分全面且准确符合要求。}\end{CJK*}\\
\end{tcolorbox}
\caption{System Prompt Used for Evaluation}
\label{fig:eval-system-prompt}
\end{figure*}

\end{document}